\documentclass[conference]{IEEEtran}

\IEEEoverridecommandlockouts
\usepackage{cite}
\usepackage{hyperref}

\usepackage{amsmath,amssymb,amsfonts}
\usepackage{graphicx}
\usepackage{textcomp}
\usepackage{xcolor}
\usepackage{algpseudocode}
\usepackage{algorithm}
\usepackage{multirow}
\usepackage{tcolorbox}
\usepackage{tikz}
\usepackage{fontawesome5}
\usepackage{ctable} 
\usepackage{booktabs}
\usepackage{ragged2e}
\usepackage{pgfplots, pgfplotstable}
\usepackage{arydshln}
\definecolor{firstcolor}{HTML}{55a868}
\definecolor{secondcolor}{HTML}{cfcfcf}
\definecolor{thirdcolor}{HTML}{c44e52}
\def\BibTeX{{\rm B\kern-.05em{\sc i\kern-.025em b}\kern-.08em
    T\kern-.1667em\lower.7ex\hbox{E}\kern-.125emX}}

\renewcommand{\baselinestretch}{0.9871}

\begin{document}

\newcommand{\insertfig}{  \includegraphics[width=\linewidth]{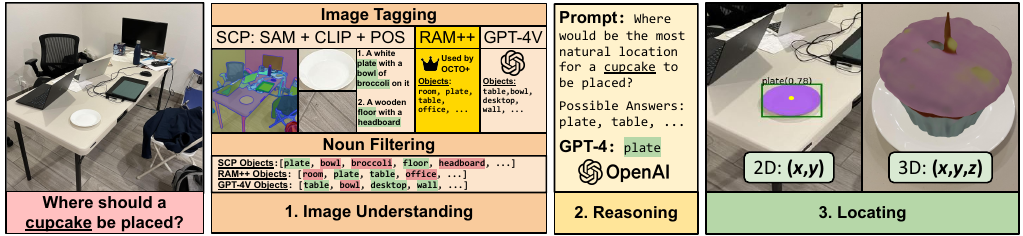}\\ \setcounter{figure}{0} \refstepcounter{figure} 
{\RaggedRight \footnotesize Fig.~\thefigure.\:\:\: Overview of various methods we experimented with, including \textbf{OCTO+} \faCrown. To determine where a \texttt{cupcake} should be placed, we perform three stages: 1) {\bf image understanding}: generate a list of all objects in the image (OCTO+ uses RAM++); 2) {\bf reasoning}: select the most natural object with GPT-4;  3) {\bf locating}: locate the 2D coordinate of the selected object in the image and ray cast the 2D coordinate to determine the 3D location in the AR scene.\vspace*{-0.9 \baselineskip}

\label{fig:teaser}}}




\makeatletter
\apptocmd{\@maketitle}{\centering \insertfig}{}{}
\makeatother

\title{OCTO+: A Suite for Automatic Open-Vocabulary Object Placement in Mixed Reality}


\author{\IEEEauthorblockN{Aditya Sharma*\thanks{*Sharma and Yoffe contributed equally to this work}, Luke Yoffe*, Tobias Höllerer}
\IEEEauthorblockA{\textit{University of California, Santa Barbara}}
Santa Barbara, California, USA \\
\texttt{\{aditya\_sharma, lukeyoffe, holl\}@cs.ucsb.edu} \\
\texttt{\color{magenta}{\href{https://octo-pearl.github.io/}{https://octo-pearl.github.io}}} 
}

\maketitle

\begin{abstract}
One key challenge in Augmented Reality is the placement of virtual content in natural locations. Most existing automated techniques can only work with a closed-vocabulary, fixed set of objects. In this paper, we introduce and evaluate several methods for automatic object placement using recent advances in open-vocabulary vision-language models. Through a multifaceted evaluation, we identify a new state-of-the-art method, OCTO+. We also introduce a benchmark for automatically evaluating the placement of virtual objects in augmented reality, alleviating the need for costly user studies. Through this, in addition to human evaluations, we find that OCTO+ places objects in a valid region over 70\% of the time, outperforming other methods on a range of metrics.

\end{abstract}


\section{\bf Introduction}


Augmented reality (AR) holds the potential to seamlessly integrate digital content into the physical world, necessitating the placement of virtual elements in natural locations. The population of 3D environments with 3D virtual content often requires developers to specify a target location, such as a ``wall", for each 3D virtual object, such as a painting. Then, while the application is running, it will recognize and track the specified location in the current scene, and anchor the virtual object on it. However, there are many cases where adding new virtual objects, including those not considered by the developers, into 3D scenes is desirable. For example, this need arises when applying custom themes for entertainment or other applications that present modified realities, such as simulation and training. Adapting AR content to different physical environments often involves placing many virtual objects, making doing this manually in every new environment cumbersome. Some automated placement techniques exist; however, their applicability is limited as they can not accommodate arbitrary objects and scenes due to the closed-vocabulary nature of the underlying machine-learning models. This constraint implies that these models can only process a predefined set of words.


On the other hand, open-vocabulary models can adapt to words not seen during training. We combine several models to create OCTO+, a state-of-the-art pipeline and evaluation methodology for virtual content placement in Mixed Reality (MR). We build on OCTOPUS~\cite{octopus}, a recently introduced 8-stage approach to the placement problem. OCTO+ accepts as input an image of a scene and a text description of a virtual object to be placed in the scene and determines the most \textit{natural} location in the scene for the object to be placed.

In summary, our contributions are three-fold:
\begin{itemize}
    \item We present the state-of-the-art pipeline OCTO+, outperforming GPT-4V and the predecessor OCTOPUS method on virtual content placement in augmented reality scenes.
    \item We conduct extensive experimentation with the state-of-the-art multimodal large language models, image editing models, and methods employing a series of models, leading to an overall 3-stage conceptualization for the automatic placement problem. 
    \item We introduce PEARL, a benchmark for \textbf{P}lacement \textbf{E}valuation of \textbf{A}ugmented \textbf{R}eality E\textbf{L}ements.
\end{itemize}

\section{\bf Related Work}
\paragraph*{\bf Virtual Content Placement}
In the context of augmented reality, virtual objects must satisfy physical and semantic constraints in order to appear \textit{natural}.

In general, virtual content should be aligned with the natural world and follow the laws of physics. For example, furniture should either be resting on the floor or against a wall~\cite{nuernberger2016snaptoreality}, and objects should not float above the ground \cite{AutoPlaceOcclude}. Evaluating such constraints automatically is a difficult task because a unique ground truth does not exist, and even if an authoritative ground truth placement were available, simply taking the distance between a proposed location and a ground truth location is not a reliable metric. The quality of a placement location also depends on many other factors, including viewing angle, viewing distance, and object size \cite{PredART}. To address this, Rafi et al.~\cite{PredART} introduced a framework that predicts how humans would rate the placement of a virtual object. The framework makes these predictions based on the ``placement gap", which is the distance between the bottom of the object and the plane it is supposed to be placed on, and other factors such as viewing angle, so it is meant to measure how physically realistic the placement of a virtual object looks. Our benchmark, on the other hand, focuses on how semantically realistic the placement of a virtual object is.

Previous work has also investigated how to place objects. To position virtual interface elements seamlessly when transitioning between physical locations, Cheng et al. \cite{SemanticAdapt} introduced an approach that discovers a semantically similar spot in the new scene corresponding to where the virtual interface elements were placed in the previous scene. To put virtual agents in AR, Lang et al.\cite{AgentPlacement} introduced a method involving reconstructing the 3D scene, identifying key objects, and optimizing a cost function based on the detected objects, among other things. Existing work focuses on placing specific objects or operates with closed-vocabulary choices. In contrast, we aim to create a single pipeline to identify \textit{any} object without special training.

\paragraph*{\bf Vision-Language Models}
Many of the models we use to automate the virtual content placement task are able to handle both image and text, but historically, machine learning models were typically either used on images or text. However, both types of models can have similar architectures, such as the Transformer architecture~\cite{transformers}. For a model to process images and text, they must be encoded into the same semantic embedding space, where similar text and images are close and unrelated text and images are distant.


One way to ``align" text and images into the same embedding space is to use Contrastive Language-Image Pretraining (CLIP)~\cite{clip}. Given $\mathcal{N}$ pairs of (text, image), a text encoder and an image encoder are used to create text embeddings $\{t_1 \dots t_{\mathcal{N}}\}$ and image embeddings $\{i_1 \dots i_{\mathcal{N}}\}$. The encoders are trained to maximize the cosine similarity between embeddings $t_a$, $i_a$ that originated from the same (text, image) pair and to minimize the cosine similarity between embeddings $t_a, i_b$ where $a \neq b$ that came from different (text, image) pairs. After this training, the encoders will map text and images into the same semantic embedding space, which allows models to take either images or text as input.
\section{\bf Method}
We explored different solutions to the overall MR object placement problem, which takes a camera frame and the natural language phrase for an object as an input and produces a 2D coordinate for the best placement of such an object as output. We frame the problem in image space for generality, as complete 3D scene models are not always available, and if they are, a 3D coordinate can easily be obtained by ray casting. While such object placement in a 2D camera frame may be a simple task for humans, it requires understanding what items are in the image, reasoning about which item in the image the object would most naturally be placed on or nearby, and finally, locating the item in the image on which the object should be placed. These subtasks are shown in \autoref{fig:teaser}, and the next section will describe each in detail.

\subsection*{\bf Stage 1. Image Understanding}
Image understanding is the process of interpreting the content in images similarly to how humans do it. In the context of object placement in augmented reality, the image understanding stage focuses on identifying all the surfaces in the image that virtual objects could be placed on. Image tagging, or recognizing all objects in an image, can be approached in different ways, including by using object-level tagging models, image-level tagging models, and multimodal large language models (LLMs).

\paragraph*{\bf Object-level Tagging}
Object-level tagging entails dividing the input image into regions of interest before generating any tags. We previously introduced the OCTOPUS pipeline, using three state-of-the-art vision and language models to accomplish this task. First, it used the Segment Anything Model (SAM)~\cite{sam} to divide the image into regions that may contain objects. Next, OCTOPUS used clip-text-decoder~\cite{cliptd} to generate captions for each region. Lastly, it used English Part-of-Speech tagging in Flair~\cite{POS} to extract nouns from each caption. We will refer to the chaining of these three models as SCP.
\paragraph* {\bf Tag Filtering}
Some nouns found may have been misidentified and must be filtered out before passing to the next stage. This is because the next stage assumes that all nouns given are represented in the image and are valid locations for an object to be placed. We experiment with multiple strategies to accomplish this.
\begin{itemize}
    \item \textbf{ViLT}~\cite{vilt}, a Vision Transformer-based model, is used in OCTOPUS for visual question answering. For every noun provided by SCP, the model is presented with the image and the question: ``Is there a {\textit{noun}} in the image?" Only the nouns that result in ViLT outputting ``yes" are retained.
    \item \textbf{CLIPSeg} ~\cite{clipseg} is a model that takes an image and text query as input and generates a heatmap illustrating the correlation between each image pixel and the text query. CLIPSeg is run on the input image and each noun, and the intensity of the brightest pixel in the resulting heatmap is recorded. This intensity is then used to rank the nouns, and the top-$k$ nouns are kept, where $k$ is a predetermined parameter.\label{clipseg}
    \item \textbf{Grounding DINO} \cite{gdino} is a model that accepts as input an image and a text query (in our case, the SCP nouns separated by commas) and outputs bounding boxes for every object it found in the image related to the query. A threshold $t$ can be adjusted to exclude boxes that are not sufficiently similar to any word in the query. We exclude any bounding boxes that cover over 90\% of the image area, as these are typically generic words such as ``room" and do not refer to specific objects in the image.
\end{itemize}

All three of these methods effectively remove nouns incorrectly identified by SCP; however, they also risk excluding too many nouns. There is a trade-off between filtering out nouns not in the image and keeping valid candidate placement targets in the image. In the case of CLIPSeg and Grounding DINO, the values of $k$ and $t$ can be adjusted to fine-tune this balance. 

\paragraph* {\bf Image-level Tagging}
It is also possible to pass the entire image into an image tagging model without dividing it into regions first. One advantage to this approach is that the model has access to the entire image, rather than just a tiny patch, and can use that to identify objects better.

The image tagging model we experimented with is the state-of-the-art Recognize Anything Plus Model (RAM++) \cite{ram++}, which takes an image and a list of text labels (this list can contain any words, including those not seen during training), and determines which of the labels are in the image. By default, the model is run on $4,585$ labels, and more can be added if needed (we found these default labels to be more than sufficient). We experimented with different thresholds for a tag to be classified as being in the image. We also tried using Grounding DINO as a tag filter since Grounding DINO was the best out of the three filter options for SCP (\autoref{table:s1}). RAM++ and Grounding DINO comprise the first stage of OCTO+.

\paragraph* {\bf Multimodal Large Language Models (MLLMs)}
Multimodal large language models can accept images and text as input, reason about them, and generate a text response. We used two such models, GPT4-V \cite{gpt4} and LLaVa-1.5 \cite{llava15}, to create a list of nouns by prompting them with the image and the following instructions:
\begin{tcolorbox}[colframe=black!1!black, title=MLLM Prompt, boxsep=0.5pt]
 \footnotesize \textbf{User Prompt:} Please provide a list of nouns (where each noun is separated by comma, example: object1, object2, object3) that could be used to describe this image. This listing should include where common indoor objects could be placed. Only list objects which are in the image. \\
 \textbf{Response:} chair, table, ladder, monitor, speaker, rack...
\end{tcolorbox}
\subsection*{\bf{Stage 2. Reasoning}}
The overall task for Stage 2 is to select the object from the list produced in Stage 1, which is the most natural location for the virtual object to be placed on. This requires complex reasoning abilities. Large language models (LLMs) have recently showcased exceptional reasoning capabilities for natural language generation. We use LLMs and multimodal LLMs to make this selection and take advantage of these abilities.
\paragraph*{\bf Large Language Model}
Both OCTOPUS and OCTO+ use OpenAI's currently most capable model, GPT-4, to select the target object on which the virtual object should be placed. Chain of Thought (CoT) prompting \cite{wei2023chainofthought} and in-context learning \cite{lampinen-etal-2022-language} are two notable techniques that increase the reasoning abilities of LLMs. As a result, we use 3-shot prompting, which means we provide 3 example questions and responses before the actual question to better guide the LLM. We provide the following prompt to GPT-4, with a temperature of $t=0.2$ (square brackets denote additional or customized information, omitted here for brevity and generality):

\begin{tcolorbox}[colframe=black!1!black, title=LLM Prompt, boxsep=0.5pt]
    \footnotesize {\bf System Prompt:} You are an expert in determining where objects should be placed in a scene. You will be given a list of objects in a scene, and the name of a new object to be placed in the scene. Your task is to select the most natural location for the new object to be placed out of the options provided. Write one of your answers and write it exactly character for character as it appears in the list of possible answers. Provide a one-word response. [Few-shot examplars] \\
    {\bf User Prompt:} \underline{Question:} Where would be the most natural location for a [object] to be placed? \underline{Possible Answers:} floor, table, computer, sink, couch 
\end{tcolorbox}

\paragraph*{\bf Multimodal Large Language Model}
The original GPT-4 model was not a multimodal LLM, so it cannot view images. Therefore, it has to rely on the list of nouns from Stage 1 to understand the content of images. By contrast, multimodal LLMs, such as GPT-4V, can take images as input and answer questions about them. As a result, MLLMs can consolidate Stages 1 and 2 into a single step.

The OCTOPUS paper experimented with visual question-answering models, such as ViLT. These models were tasked with processing an image and a text prompt that inquired about the optimal placement of an object within the image. At that time, the models frequently produced unsatisfactory responses. However, significant progress has since occurred in the field, so our experimentation now includes state-of-the-art closed-source and open-source models, specifically GPT-4V (via the API)~\cite{openai2023gpt4} and LLaVA-v1.5-13B (temperature $t=0.01$)~\cite{liu2023improved}. We directly provide the models with the image and object to be placed, prompting them to name where in the image the object should be placed:
\begin{tcolorbox}[colframe=black!1!black, title=MLLM Prompt, boxsep=0.5pt]
\footnotesize
{\bf System Prompt:} You are an expert in determining where objects should be placed in a scene. You are given an image of a scene the name of an object to be placed in the scene. Please respond with a concise answer, one or two words, naming the object in the scene on which the new object should be placed.\\
{\bf User Prompt:} Where would be the most natural location to place a [object] in this image?
\end{tcolorbox}

\subsection*{\bf{Stage 3: Locating}} \label{sec:Target Locating Methods}
Once a suitable surface for the virtual object's placement has been found, the next task is to select a 2D coordinate for the object to be placed. We explore two distinct strategies:

\begin{enumerate}
    \item {\bf CLIPSeg} As discussed in the \hyperref[clipseg]{Tag Filtering} section, CLIPSeg generates a heatmap indicating the similarity between each pixel in the image and the provided text query. We present CLIPSeg with our image and surface selected in Stage 2 to determine the object's placement. We choose the location $(x, y)$ of the pixel with the highest activation in the heatmap. This is the model used by OCTOPUS.
    \item {\bf Grounded-Segment-Anything} \cite{gsam} is another approach that integrates Grounding DINO and SAM. As input, we provide the image and text specifying the surface chosen in Stage 2. First, Grounding DINO identifies bounding box(es) corresponding to the text. These boxes are input to SAM, which generates a precise segmentation mask around the object. The masks are consolidated into one, and we select the point in the combined mask that is farthest away from any edge of the mask. This ensures that we do not select a point right at the edge of a surface, which would not be very natural. This is the model used by OCTO+.\end{enumerate}

We also examine performing Stages 1-3 in one step using GPT-4V. We ask GPT-4V to determine the $(x,y)$ pixel location directly with the following prompt: 
\begin{tcolorbox}[colframe=black!1!black, title=GPT4-V Prompt, boxsep=0.5pt]
\footnotesize
{\bf System Prompt:} You are an expert in determining where objects should be placed in a scene. You are given an image of a scene and the name of an object to be placed in the scene. Please respond with the pixel coordinates locating where in the scene would be the most natural location to place the object. The image is 561 pixels wide and 427 pixels long, and the top left corner is the origin.\\
{\bf Chat Prompt:} Where would be the most natural location to place a [object] in this image? Please briefly explain your selection and enter the x and y coordinate locations in the format (x,y) at the end. Example: The banana should be placed on the table. (173, 294).
\end{tcolorbox}

One final approach to obtain the 2D coordinate for object placement in an image starts with InstructPix2Pix~\cite{brooks2023instructpix2pix}. InstructPix2Pix is an instruction-based image editing model that takes both an image and a text prompt specifying the desired edit. Using Stable Diffusion, InstructPix2Pix then generates a new image incorporating the edit. In our case, where we seek to identify the optimal placement of an object, we prompt InstructPix2Pix to ``add [object]" and provide the input image. Leveraging its image understanding and reasoning capabilities, InstructPix2Pix generates a new image with the object seamlessly integrated. We then use Grounding-Segment-Anything to detect the object and segment the region. To determine the 2D placement coordinate, we select the bottom-most pixel in the mask, considering it to be the location of the surface beneath the object.

\subsection*{\bf 3D Location in AR Scene}
Once the 2D ($x, y$) location in the image is identified, the final step is to calculate the corresponding 3D ($x, y, z$) position in the scene, which is where the virtual object will be positioned in augmented reality (AR). To accomplish this, we employ raycasting into the scene (executed by ARKit and ARCore, supported natively in iOS and Android devices).

\begin{figure}[t]
    \centering
    \includegraphics[width=0.35\textwidth]{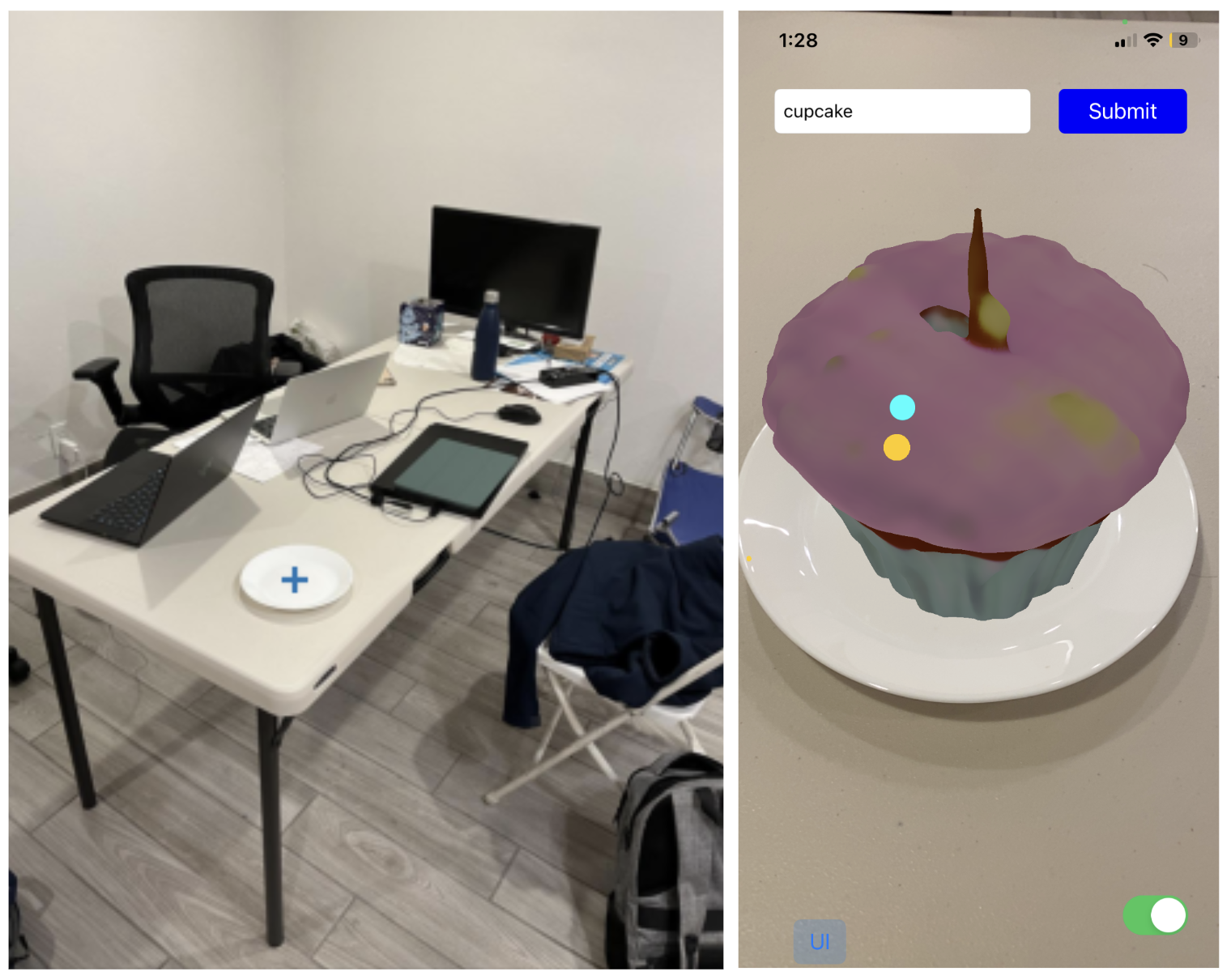}
    \caption{\textbf{Left:} The 2D location in the image selected. \textbf{Right:} A screenshot of the 3D scene with a virtual cupcake placed on the plate. Both the 2D and 3D locations were found as described in the \hyperref[sec:Target Locating Methods]{Locating} section.}
    \label{fig:2d-and-3d-locs}
    \vspace*{-3ex}
\end{figure}

\section{\bf Evaluation and Results}
To determine which models performed the best in each stage, we designed a series of experiments, one for each stage, to compare the different methods.

\paragraph*{\bf Experiment Setup} A representative set of objects and images is needed for meaningful and fair evaluation. We limited our assessment to indoor scenes. By consulting LLMs and optimizing for object diversity, we compiled a list of 15 common indoor objects: \texttt{(apple, cake, cup, plate, vase, stool, painting, lamp, book, bag, computer, pencil, shoes, cushion, cat)}. This step was decoupled entirely from the scene selection, for which we randomly sampled 100 indoor scene images from the NYU Depth Dataset~\cite{nyudepth} and Sun3D Dataset~\cite{sun3d}.

\begin{figure}[t]
    \centering
    \includegraphics[width=\columnwidth]{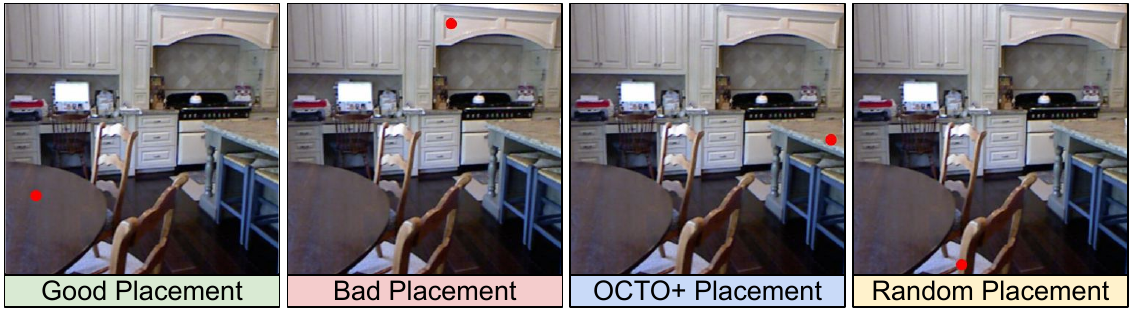}
    \vspace*{-3ex}
    \caption{Comparative placements for one input image, and the prompt ``Apple".}
    \label{fig:placement_imgs}
    \vspace*{-2.5ex}
\end{figure}

\paragraph*{\bf Annotation} We had two team members annotate each of the 100 images with a {\em natural} and {\em unnatural} location to place each of the 15 objects. For example, it would be {\em natural} to place a cupcake on a plate, but {\em unnatural} to place a cupcake on the floor. Team members also wrote a list of valid locations for each object to be placed in each image. \autoref{fig:placement_imgs} describes the different placements. Any objects that were deemed unsuitable or irrelevant for a specific image were excluded from further analysis throughout the experiment for that particular image (this happened in 573 of the 1,500 image-object combinations). We also generated placement coordinates using random point selection. Ultimately, we arrived at 927 object location-image pairs for each of the three baseline placement methods. These images and lists of valid locations for each object in each image annotation are a part of the PEARL benchmark and are used to evaluate all three stages.

\subsection*{\bf Stage 1: Image Understanding Evaluation}
Stage 1 must produce an accurate list of objects in the image for the later stages to succeed. This means that all the important objects in the image should be included in the list, and objects not in the image should not. We consider objects important if they are surfaces on which other objects could be placed, such as tables, chairs, kitchen counters, or the floor. In other words, the necessary objects are the valid locations annotated as described above. Therefore, we compare the list of objects generated in Stage 1 with the valid locations in two ways:
\begin{itemize}
    \item \textbf{Exact Match (EM):} For each object (e.g., ``cat") in each image, we check if any of its valid locations (e.g., ``couch," ``floor") is in the list of tags
    \item \textbf{Sentence-BERT} \cite{sbert} (sBERT) Similarity: For each object (e.g., ``cat") in each image, we find the maximum similarity between any of its valid locations (e.g., ``couch," ``floor") and any of the tags. We do this using sBERT, a modified version of BERT \cite{bert} that encodes text into a semantic embedding space, such that the cosine similarity between the embedding for words that have similar meanings (e.g., sBERT(``couch", ``sofa") = 0.856) will be higher than it would be for words that have different meanings (e.g., sBERT(``couch", ``table") = 0.334). Using sBERT, we can reward methods that generate tags with the correct semantic meaning.
\end{itemize}

We used these techniques to compute overall metrics for the tagging stage using the algorithm shown in Algorithm \autoref{alg:s1}, which computes the average score for each image, and then averages each image's score to produce a final score.  In addition to the average number of tags generated per image, these metrics are shown for each method in \autoref{table:s1}.

Starting with the object-level tagging method, we found that the unfiltered list of tags was very long (60.67 on average), with many of the nouns being irrelevant or not in the image. Out of the filtration methods we tried, we felt that Grounding DINO with a threshold of 0.25 had the best trade-off between keeping relevant nouns (the number of EMs and sBERT score only dropped by a few percent compared to unfiltered), while the average number of tags per image dropped by nearly 75\%. Taking the top-20 tags from CLIPSeg also performed well, so we experimented with both of these in the later stages. A trade-off exists where higher thresholds result in removing more undesired words and excluding more target words. We generally leaned towards less strict thresholds, as excluding a target word is usually more harmful than including an extra word.

On to the image-level tagging results, we found that RAM++ with 0.8 times the default threshold, combined with the best filtering strategy from before (Grounding DINO with a threshold of 0.25) had the best performance of any Stage 1 method. Compared to RAM++ with the default threshold and no filter, this method had much better metrics and only a slightly longer average number of tags.

Lastly, we considered the multimodal large language models, which were prompted to generate a list of nouns in the image. We frequently omitted crucial nouns in both models, so we did not experiment further with their noun lists. Additionally, LLaVA often repeats sequences of words repeatedly, making it impractically slow. This was not an issue with GPT4-V, however.

\begin{algorithm}[h]
\caption{State 1 Metrics Computation}\label{alg:s1}

\begin{algorithmic}
\Require $\mathcal{I}$: 100 images
\Require $\mathcal{T}$: model-generated tags generated for each image
\Require $\mathcal{O}$: names of up to 15 objects placed in each image
\Require $\mathcal{L}$: valid locations for each object in each image
\State $\mathcal{E} \gets 0$ \Comment{Number of exact matches}
\State $\mathcal{S} \gets 0$ \Comment{Total sBERT score}
\State{$\mathcal{N} \gets length(\mathcal{I})$} \Comment{Number of Tags}
\For{$i \in 1$ to $\mathcal{N}$}
    \State $e \gets 0$
    \State $s \gets 0$
    \State{$\mathcal{M} \gets length(\mathcal{O}_i)$}
    \For{$j \in 1$ to $\mathcal{M}$}        
        \If{$\mathcal{T}_i \cap \mathcal{L}_{i,j} \neq \emptyset$}
            \State $e \gets e + 1$
        \EndIf
        \State $s \gets s + \max_{(t, l) \in \mathcal{T}_i \times \mathcal{L}_{i,j}} \text{sBERT}(t,l)$        
    \EndFor
    \State $\mathcal{E} \gets \mathcal{E} + \frac{e}{\mathcal{M}}$
    \State $\mathcal{S} \gets \mathcal{S} + \frac{s}{\mathcal{M}}$
\EndFor
\State $\mathcal{E} \gets \frac{\mathcal{E}}{\mathcal{N}}$
\State $\mathcal{S} \gets \frac{\mathcal{S}}{\mathcal{N}}$
\end{algorithmic}
\end{algorithm}

\begin{table}[]
  \centering
  \begin{tabular*}{\columnwidth}{@{\extracolsep{\fill}}lccc}
  \specialrule{.10em}{.001em}{.001em}
  \textbf{Models} & 
  \textbf{Exact Match} \faArrowUp[regular]
  & \textbf{sBERT} \faArrowUp[regular] & \textbf{\# Tags} \faArrowDown[regular] \\
  \specialrule{.075em}{.001em}{.001em}
  \multicolumn{4}{c}{\textbf{Object-level Tagging}} \\
  \specialrule{.075em}{.001em}{.001em}
  SCP (SAM + CLIP + POS)          & \textbf{0.734}    & \textbf{0.892}    & 60.67  \\
  ~~ + ViLT             & 0.658             & 0.850             & 25.96 \\
  ~~ + CLIPSeg ($k$=10) & 0.439             & 0.698             & \underline{10.00} \\
  ~~ + CLIPSeg ($k$=15) & 0.535             & 0.755             & 15.00 \\
  ~~ + CLIPSeg ($k$=20) & 0.634             & 0.807             & 20.00 \\
  ~~ + CLIPSeg ($k$=30) & 0.684             & 0.841             & 29.95 \\
  ~~ + G-DINO ($t$=0.25)  & \underline{0.712} & \underline{0.852} & 16.04 \\
  ~~ + G-DINO ($t$=0.35)  & 0.532             & 0.745             & \textbf{7.87} \\
  \specialrule{.075em}{.001em}{.001em}
  \multicolumn{4}{c}{\textbf{Image-level Tagging}} \\
  \specialrule{.075em}{.001em}{.001em}
  RAM++ ($t$=0.8) & \textbf{0.909} & \textbf{0.976} & 61.66 \\ 
  ~~~ + G-DINO ($t$=0.25) & \underline{0.865} & \underline{0.942} & 18.51 \\
  RAM++ ($t$=0.9) & 0.851 & 0.937 & 31.63 \\ 
  ~~~ + G-DINO ($t$=0.25) & 0.812 & 0.923 & \underline{15.67} \\
  RAM++ ($t$=1.0) & 0.619 & 0.854 & 16.55 \\
  ~~~ + G-DINO ($t$=0.25) & 0.610 & 0.834 & \textbf{10.84} \\
  \specialrule{.075em}{.001em}{.001em}
  \multicolumn{4}{c}{\textbf{Multimodal Large Language Models}} \\
  \specialrule{.075em}{.001em}{.001em}
  GPT-4V(ision) & 0.399 & 0.782 & \textbf{11.69} \\
  LLaVA-v1.5-13B & \textbf{0.605} & \textbf{0.805} & 29.77 \\
  \specialrule{.10em}{.001em}{.001em} \\
  \end{tabular*}
    \caption{Stage 1 Metrics. Best \textbf{IN BOLD}, second best \underline{UNDERLINED}.}
    \label{table:s1}
\vspace*{-6ex}
\end{table}

\begin{table}[]
  \centering
  \begin{tabular*}{\columnwidth}{@{\extracolsep{\fill}}lcc}
    \specialrule{.10em}{.001em}{.001em}
    \textbf{Models} & \textbf{Exact Match \faArrowUp[regular]} & \textbf{sBERT \faArrowUp[regular]} \\
    \specialrule{.075em}{.001em}{.001em}
    \multicolumn{3}{c}{\textbf{LLM: Tags + Object as Input}} \\
    \specialrule{.075em}{.001em}{.001em}
    SCP (Filter: CLIPSeg) + GPT-4 & 0.621 & 0.786 \\
    SCP (Filter: G-DINO) + GPT-4 & \underline{0.630} & \underline{0.791} \\
    RAM++ (Filter: G-DINO) + GPT-4 & \textbf{0.645} & \textbf{0.821} \\
    \specialrule{.075em}{.001em}{.001em}
    \multicolumn{3}{c}{\textbf{Multimodal LLMs: Image + Object as Input}}\\
    \specialrule{.075em}{.001em}{.001em}
    GPT-4V(ision) & \underline{0.479} & \underline{0.754} \\
    LLaVA-v1.5-13B & \textbf{0.711} & \textbf{0.844} \\
    \specialrule{.10em}{.001em}{.001em} \\
  \end{tabular*}
  \caption{Stage 2 Metrics. Best \textbf{IN BOLD}, second best \underline{UNDERLINED}}
  \label{table:selector}
  \vspace*{-6ex}
\end{table}

\subsection*{\bf Stage 2: Reasoning Evaluation}
We measure the performance of this stage on the PEARL benchmark by comparing the target tag $\hat{\mathcal{T}}$ produced a method with the set of expert annotated tags $\mathcal{T}$ for each object in each image. To measure the performance, we use two metrics:
\begin{itemize}
    \item \textbf{Exact Match:} For each object in each image, we check if the predicted surface $\hat{\mathcal{T}}$ matches any of the expert annotated tags in $\mathcal{T}$
    \item \textbf{sBERT:} To reduce how much we penalize methods that select a word with the correct meaning but not an exact match (e.g., ``couch" vs. ``sofa"), we record the maximum similarity between $\hat{\mathcal{T}}$ and any of the annotated tags in $\mathcal{T}$.
\end{itemize}

We compute the average exact match and sBERT score over each object in each image and then average over all images to get a final score (there are 927 image-object pairs). We report the scores for the five methods we tried in \autoref{table:selector}. We find that the SCP tags filtered with G-DINO perform better than the SCP filtered by CLIPSeg with EM scores of 0.630 and 0.621, respectively. Still, RAM++ outperforms SCP with an EM of 0.645 and an sBERT score of 0.821. In Multimodal LLMs, we find that the open-source LLaVA-v1.5-13B model significantly outperforms GPT-4V. Although this is the case, we find that LLaVA is substantially slower (when run on an NVIDIA TITAN X Pascal GPU) than GPT-4V, making it impractical to be used in a real-time AR application. Furthermore, in \autoref{table:locator}, it is evident that GPT-4V outperforms LLaVA with a higher score, suggesting that Stage 2 metrics may not fully represent the overall task.


\subsection*{\bf Stage 3: Target Locating Evaluation}
Measuring the performance of Stage 3 is the most important as it is the only way to evaluate the system end-to-end. Even if the first two stages correctly select the target noun, determining a \textit{natural} location for placement is still required. Therefore, we establish a robust evaluation procedure incorporating automated metrics and human evaluation.

\begin{table*}[t]
  \centering
  \begin{tabular*}{\textwidth}{@{\extracolsep{\fill}}llllcccc}
    \specialrule{.10em}{.001em}{.001em}
    \multicolumn{4}{c}{\textbf{Pipeline \faLayerGroup[regular]}} & \multicolumn{2}{c}{\textbf{Automated Metrics} \faChartBar[regular]} & \multicolumn{2}{c}{\textbf{Human Evaluation} \faUser}  \\
    \textbf{Tagger \faTags} & \textbf{Filter \faFilter} & \textbf{Selector \faCheck} & \textbf{Locator \faMapMarker} & \textbf{\textsc{In Mask} \faArrowUp[regular]} & \textbf{\textsc{Score \faArrowUp[regular]}} & \textbf{\textsc{MTurk \faArrowUp[regular]}} & \textbf{\textsc{Expert \faArrowUp[regular]}}\\
    \specialrule{.10em}{.001em}{.001em}
    \multicolumn{8}{c}{\textbf{Baselines}} \\
    \specialrule{.075em}{.001em}{.001em}
         --------- & --------- & ---------  & Natural Placement  & \textbf{0.907} & \textbf{17.987} & \textbf{1.000} & \textbf{1.000} \\
     --------- & --------- & ---------& Random Placement & \underline{0.161} & \underline{-106.113} & \underline{0.467} & \underline{0.040} \\
     --------- & --------- & ---------& Unnatural Placement & 0.010 & -176.375 & 0.167 & 0.020 \\
    \specialrule{.075em}{.001em}{.001em}
    \multicolumn{8}{c}{\textbf{Selected Tag + Image as Input}} \\
    \specialrule{.075em}{.001em}{.001em}
    SCP & CLIPSeg & GPT-4 & CLIPSeg (Max) & 0.572 & -20.730 & --------- & --------- \\
    \faCertificate \ SCP & ViLT & GPT-4 & CLIPSeg (Max) & 0.588 & -15.300 & 0.514 & 0.570 \\
    SCP & G-DINO & GPT-4 & CLIPSeg (Max) & 0.596 & -13.005 & --------- & --------- \\
    SCP & CLIPSeg & GPT-4 & G-SAM (Center) & 0.613 & -10.783 & --------- & --------- \\
    SCP & G-DINO & GPT-4 & G-SAM (Center) & 0.615 & -6.464 & --------- & --------- \\
    --------- & --------- & LLaVa-v1.5-13B & CLIPSeg (Max) & 0.649 & -13.17 & --------- & --------- \\
    RAM++ & G-DINO & GPT-4 & CLIPSeg (Max) & 0.671 & -4.185 & 0.547 & --------- \\
    --------- & --------- & GPT-4V(ision) & G-SAM (Center) & 0.686 & \underline{4.317} & \underline{0.580} & --------- \\
    --------- & --------- & GPT-4V(ision) & CLIPSeg (Max) & \underline{0.692} & -4.492 & \textbf{0.582} & \underline{0.620}\\
    \hdashline[4pt/4pt]
    \faCrown \ RAM++ & G-DINO & GPT-4 & G-SAM (Center) & \textbf{0.702} & \textbf{7.634} & 0.527 & \textbf{0.690} \\
    \specialrule{.075em}{.001em}{.001em}
    \multicolumn{8}{c}{\textbf{Object + Image as Input}} \\
    \specialrule{.075em}{.001em}{.001em}
    --------- & --------- & InstructPix2Pix & G-SAM (Bottom) & \underline{0.283} & \underline{-60.852} & --------- & --------- \\
    --------- & --------- & --------- & GPT-4V (Pixel Location) & \textbf{0.321} & \textbf{-34.282} & --------- & --------- \\
    \specialrule{.1em}{.001em}{.001em}
  \end{tabular*}
  \caption{Stage 3 Metrics. Best \textbf{IN BOLD}, second best \underline{UNDERLINED}. \faCrown \ is OCTO+, \faCertificate \ is OCTOPUS}
  \label{table:locator}
  \vspace*{-5ex}
\end{table*}

\paragraph*{\bf Automated Metrics}
The 98 indoor scene images used in the PEARL benchmark came from the NYU Depth Dataset and Sun3D dataset, which included segmentation masks for many indoor objects. Some of the images did not come with any segmentation masks labeled with any of PEARL's annotated ground truth locations for particular objects (e.g., ``desk" and ``table" segmentation masks were not provided in an image for the indoor classroom, so we were not able to evaluate how well an apple could be placed in that image). As a result, we had to remove 152 of the image-object pairs from the original 927, leaving us with 775 image-object pairs for which we can compute metrics. Each pixel in the $561\times427$ segmentation mask was annotated with a value between 1 and 895, where each value was mapped to a unique label.

To evaluate how naturally an object is placed in an image, PEARL combines all of the segmentation masks whose label matches one of the ground truth locations for that object in that image. For instance, a ``cat" could be naturally placed on a chair, floor, or couch in a given image. To obtain a new segmentation mask, assigning a value of 1 in all valid locations for the cat and 0 elsewhere, we consolidate the segmentation masks for that image corresponding to the labels chair, floor, or couch. In some cases, the segmentation masks had more specific labels than PEARL or used synonyms, so we had to change some of the labels in the segmentation dataset manually. For example, we changed ``sofa" to ``couch" and merged ``coffee table" and ``desk" into ``table." In total, we modified 43 segmentation mask labels.



In the following descriptions of our metrics, we will use ($\hat{x}_{ij},\hat{y}_{ij}$) to denote the 2D location that a model output for object $j$ in image $i$. When we refer to a ``mask" for an object and image, we refer to the set of valid locations for that object to be placed in that image.

The first automated metric we define is the percentage of the predicted 2D coordinate ($\hat{x}_{ij},\hat{y}_{ij}$) was in the mask. For a given placement ($\hat{x}_{ij},\hat{y}_{ij}$) of object $j$ in image $i$, we assign a score of 1 if ($\hat{x}_{ij},\hat{y}_{ij}$) is in the mask, and 0 otherwise. We repeat this for every object in an image and find the proportion of times the predicted location was in the mask. We then take the average of all images' scores to compute the final \textsc{In Mask} score in \autoref{table:locator}.


The second automated metric we define is PEARL-Score. We observed that the \textsc{In Mask} score does not consider the distance between the point and the mask. In other words, if the predicted location is not in the mask, it is treated the same as if the point was on the other side of the image. In addition, the score does not consider where the object is to be placed in the mask. To humans, it is more \textit{natural} to place objects away from the edge of a surface. For example, one would not place a ``cup" on the corner of a table; rather, they would place it where the ``cup" is more centered and not at risk of falling. Therefore, we designed PEARL-Score to be negative when the placement location is outside the mask, and positive when it is inside the mask. We give more points if the placement is well within the mask, and subtract more points if the placement is far outside the mask.
We divide this into three cases: 
\begin{enumerate}
    \item Case 1: If the 2D Coordinate ($\hat{x}_{ij},\hat{y}_{ij}$) is in the mask, we find the closest point $(x, y)$ outside the mask. Then, the PEARL-Score is the Euclidean distance between the two points $\mathcal{L}_2(x, y, \hat{x}, \hat{y})$ as defined in \autoref{eq:3}. The farther away ($\hat{x}_{ij},\hat{y}_{ij}$) is from the edge of the mask, the higher the PEARL-Score. A point near the edge of the mask would have a lower PEARL-Score.
    \item Case 2: If the 2D Coordinate ($\hat{x}_{ij},\hat{y}_{ij}$) is outside the mask, we find the closest point $(x, y)$ inside the mask. Then, the PEARL-Score would be $-\mathcal{L}_2(x, y, \hat{x}, \hat{y})$. The score is negative because the placement should be penalized for being outside the mask.
    \item Case 3: If the 2D Coordinate ($\hat{x}_{ij},\hat{y}_{ij}$) is one the edge of the mask, then the PEARL-Score would be 0.
    
\end{enumerate}
The PEARL-Score can be summarized in \autoref{eq:1}, where:
\begin{itemize}
    \item $S$ is the overall PEARL-Score on a set of $\mathcal{N}$ images.
    \item $\mathcal{M}_i$ is the number of objects being placed in image $i$.
    \item $\mathcal{V}_{ij}(x,y)$ is the value of the mask of valid locations for object $j$ in image $i$ at point $(x,y)$. Its value is 0 outside the mask and 1 inside the mask.
    \item $(\hat{x}_{ij}, \hat{y}_{ij})$ is the predicted placement location for object $j$ in image $i$.
\end{itemize}

\begin{equation}
    \mathcal{S} = \frac{1}{\mathcal{N}}\sum_{i=1}^{\mathcal{N}} \frac{1}{\mathcal{M}_i} \sum_{j=1}^{\mathcal{M}_i} (-1)^{1 - \mathcal{V}_{ij}(\hat{x}_{ij}, \hat{y}_{ij})} \cdot \mathcal{D}(\hat{x}_{ij}, \hat{y}_{ij})
    \label{eq:1}
\end{equation}
\begin{equation}
    \mathcal{D}(\hat{x}_{ij}, \hat{y}_{ij}) = \min_{\substack{x, y : V_{ij}(x, y) = \\1 - V_{ij}(\hat{x}_{ij}, \hat{y}_{ij})}} \mathcal{L}_2(x, y, \hat{x}_{ij}, \hat{y}_{ij})
    \label{eq:2}
\end{equation}
\begin{equation}
    \mathcal{L}_2(x, y, \hat{x}_{ij}, \hat{y}_{ij}) = \sqrt{(\hat{x}_{ij} - x)^2 + (\hat{y}_{ij} - y)^2}
    \label{eq:3}
\end{equation}
$\mathcal{D}(\hat{x}_{ij}, \hat{y}_{ij})$, defined in \autoref{eq:2}, is the minimum distance between the predicted placement location and a point on the edge of the mask. The score for an image is calculated by averaging the scores of all object placements within the image. Finally, the overall PEARL-Score $\mathcal{S}$ is computed by taking the average score across all images.



\paragraph*{\bf Human Evaluation}
To verify the agreement of our automated metrics with human preferences, we supplement our results with a human evaluation. To gather opinions from a general audience, we conducted an Amazon Mechanical Turk (MTurk) study to evaluate OCTO+, OCTOPUS, and three of the other placement techniques we implemented that performed the best on our automated metrics:
\begin{itemize}
    \item \textbf{OCTO+:} \faCrown\hspace{0.1cm} RAM++ (G-DINO) + GPT-4 + G-DINO
    \item \textbf{OCTOPUS:} \faCertificate\hspace{0.1cm} SCP (ViLT) + GPT-4 + CLIPSeg
    \item RAM++ (G-DINO) + GPT-4 + CLIPSeg
    \item GPT-4V + G-DINO
    \item GPT-4V + CLIPSeg
\end{itemize}

To evaluate each placement method, we used them to generate 2D placements for 100 randomly selected image-object pairs. We divided the 100 images into Human Intelligence Tasks (HITs), each containing five images, for a total of 20 HITs for each of the evaluations listed above. For the baselines (expert \textit{unnatural} and random), we only ran the assessment on 50 images, or 10 HITs.

In each HIT, evaluators were told what object was to be placed and were shown two images side-by-side. Both images were annotated with a red circle indicating a proposed placement location. One of the proposed placement locations came from the evaluated method, while the other was a \textit{natural} placement annotated by an expert. The evaluators then selected which placement location was superior or declared a tie if both locations were deemed equally appropriate for the object. The evaluators did not know which method produced each placement location.

To ensure our data was of high quality, we specified the following criteria workers must meet to work on the task:
\begin{itemize}
    \item Only allow workers with 95\%+ HIT Approval Rate
   \item Only allow workers with 50+ HITs approved
   \item Only allow workers from regions in US \& UK
   \item Added a sixth side-by-side comparison of an obvious good vs bad placement as an ``attention check".
\end{itemize}

In \autoref{table:locator}, the results of the Mechanical Turk study are shown in the column labeled \textsc{MTurk}. Suppose a placement is tied with a \textit{natural} placement. In that case, it must be natural, so to compute the metrics shown in the table, we added the proportion of the time that the method in question won or tied against the \textit{natural} placement.

The results showed that GPT-4V with CLIPSeg as the locator performed best, tying or winning against the expert \textit{natural} placements 58.2\% of the time. GPT-4V with G-SAM was second place, followed by the RAM++ variants (which include OCTO+), with OCTOPUS (SCP + ViLT + GPT-4) performing the worst out of the methods that we ran an evaluation for.

Looking at our baselines, the expert unnatural placements won or tied with the expert natural placements 16.7\% of the time, which is higher than expected and could indicate data noise. We also performed an expert-level human evaluation to mitigate noise in the data. Two of our team members evaluated OCTOPUS, OCTO+, the best performing GPT-4V method, and the baselines (random and \textit{unnatural}). The format mirrored the MTurk study, displaying evaluators two side-by-side images: one generated using the evaluated method and the other annotated by experts with \textit{natural} placement. This process was performed on 50 randomly selected object-image pairs for baselines and 100 for other comparisons.

The results show in \textsc{Expert} column of \autoref{table:locator}, reveal that in the judgment of the two evaluators, 69\% of the time, OCTO+ selected a location at least as natural as the human expert selecting a {\em natural} location. Comparing this with GPT4-V, only 62\% of the time and SCP only 57\% of the time did the human experts select a natural location. The experts' natural locations won over the random and unnatural locations 96\% and 98\% of the time, respectively, which confirms that they were indeed appropriate locations the vast majority of the time, demonstrating that it is tailored to human preferences and not far off from the expert's {\em natural} placements. 

In summary, as documented by \autoref{table:locator}, OCTO+  performs the best in three of the four evaluation studies evaluated and remains competitive in the fourth evaluation study.

\section{\bf Discussion and Limitations}

While our method generally places objects naturally, it has limitations.  First, it takes up to 10 seconds to generate a single placement location on an NVIDIA RTX A4000, which means it is not yet truly practical in real-world applications, in particular when making live queries with AR cameras. We specifically tested it in such a use case (see Section \ref{sec:application} below), and the latency may pose an inconvenience to users. Additionally, while our method excels at selecting the optimal entity for virtual object placement and aims for a centrally located point on the surface, the outcome is not always the most \textit{natural}. For example, it is conventional for people to hang paintings at eye level, a consideration our pipeline currently lacks. Our pipeline can also struggle when the surface selected is a complex shape. For example, if our pipeline determines that a cat should be placed on a couch, it will not consider that the cat would most naturally be placed on the seat, and may select a location on the backrest or side instead. This could potentially be addressed with complete consideration of the 3D model of the scene, which would enable us to restrict placement locations to horizontal or vertical planes, depending on the object being placed. Enhancing 3D reasoning with vision-and-language models will eventually yield even better results.  

\section{\bf Application}
\label{sec:application}
To demonstrate the practical purposes of our model, we created an iOS AR application that uses OCTO+, as shown in the right image of \autoref{fig:2d-and-3d-locs}. The app takes a text prompt as input to convert this text prompt into a 3D model and then places the model in the scene at a natural location. For example, if the text input was ``\texttt{a cupcake}" and the real-world scene contained a pair of shoes, a backpack, and a table with a plate on it, then our app would generate a 3D model of a red cake and place it on the plate. We use ARKit\cite{arkit} to track the device and identify planes in the image on which rays can be cast and SceneKit\cite{scenekit} to render 3D models. We also have a backend where two steps are performed offline:

\paragraph{Object Generation}
To generate a 3D model of the text prompt, we use OpenAI's Shap-E\cite{shapE}, a text-to-3D diffusion model. Shap-E can take in any text as input, so it is not restricted to a specific list of objects like a library of 3D assets would be. Among text-to-3D models, we choose Shap-E because it is fast (20 seconds on an NVIDIA RTX A4000) and generates acceptable-quality models.

\paragraph{Object Placement}
Our pipeline takes the camera frame from when the user submits their input text prompt and returns the 3D world coordinates to place the object. Once the object is placed in the scene, the user can see it and look around it. Multiple objects can be generated to give a room a theme. For example, to create a Halloween-themed room, the user could create pumpkins, skeletons, and candy. This AR experience can enable users to design their own themed virtual environments.

\section{\bf Conclusion}
We present OCTO+, a state-of-the-art method for placing virtual content in augmented reality. OCTO+ outperforms its successor OCTOPUS and the state-of-the-art multimodal model GPT-4V in three of the four metrics observed in this paper. The OCTO+ pipeline is built using RAM++ as the image-tagging model, G-DINO as the filter, GPT-4 LLM as the reasoner to select the best object in the image, and G-SAM as the locator to choose the more natural 2D location. The entire OCTO+ pipeline is open-vocabulary, meaning it can be used to place \textit{any} object in \textit{any} scene out of the box without \textit{any} fine-tuning. We also present PEARL-Score, an automated metric aligned to human preferences. PEARL introduces a challenging benchmark in virtual content placement in augmented reality.

In future work, we would like to accelerate further the placement determination (which currently takes up to 10s) to enable truly interactive Mixed Reality experiences. Also, we are exploring relaxations of the placement phrasing, which presently only considers placing objects {\em on} elements visible in the picture. Other prepositions (e.g., ``above" and ``near") could be considered. Object orientation could also be specifically specified and addressed (e.g. ``facing the camera/facing the window"). LLMs can adeptly manage complex, vague, and multi-object spatial directives (e.g., ``add paintings and poster to this room" or ``add Christmas decorations"). Future research can extend their capabilities in handling such directives.

\bibliographystyle{./IEEEtran}
\bibliography{./IEEE}

\begin{thebibliography}{10}
\providecommand{\url}[1]{#1}
\csname url@samestyle\endcsname
\providecommand{\newblock}{\relax}
\providecommand{\bibinfo}[2]{#2}
\providecommand{\BIBentrySTDinterwordspacing}{\spaceskip=0pt\relax}
\providecommand{\BIBentryALTinterwordstretchfactor}{4}
\providecommand{\BIBentryALTinterwordspacing}{\spaceskip=\fontdimen2\font plus
\BIBentryALTinterwordstretchfactor\fontdimen3\font minus \fontdimen4\font\relax}
\providecommand{\BIBforeignlanguage}[2]{{%
\expandafter\ifx\csname l@#1\endcsname\relax
\typeout{** WARNING: IEEEtran.bst: No hyphenation pattern has been}%
\typeout{** loaded for the language `#1'. Using the pattern for}%
\typeout{** the default language instead.}%
\else
\language=\csname l@#1\endcsname
\fi
#2}}
\providecommand{\BIBdecl}{\relax}
\BIBdecl

\bibitem{octopus}
L.~Yoffe, A.~Sharma, and T.~H{\"o}llerer, ``Octopus: Open-vocabulary content tracking and object placement using semantic understanding in mixed reality,'' in \emph{2023 IEEE International Symposium on Mixed and Augmented Reality Adjunct (ISMAR-Adjunct)}.\hskip 1em plus 0.5em minus 0.4em\relax IEEE, 2023, pp. 587--588.

\bibitem{nuernberger2016snaptoreality}
B.~Nuernberger, E.~Ofek, H.~Benko, and A.~D. Wilson, ``Snaptoreality: Aligning augmented reality to the real world,'' in \emph{Proceedings of the 2016 CHI Conference on Human Factors in Computing Systems}, 2016, pp. 1233--1244.

\bibitem{AutoPlaceOcclude}
D.~E. Breen, R.~T. Whitaker, E.~Rose, and M.~Tuceryan, ``Interactive occlusion and automatic object placement for augmented reality,'' in \emph{Computer Graphics Forum}, vol.~15, no.~3.\hskip 1em plus 0.5em minus 0.4em\relax Wiley Online Library, 1996, pp. 11--22.

\bibitem{PredART}
T.~Rafi, X.~Zhang, and X.~Wang, ``Predart: Towards automatic oracle prediction of object placements in augmented reality testing,'' in \emph{37th IEEE/ACM International Conference on Automated Software Engineering}, 2022, pp. 1--13.

\bibitem{SemanticAdapt}
Y.~Cheng, Y.~Yan, X.~Yi, Y.~Shi, and D.~Lindlbauer, ``Semanticadapt: Optimization-based adaptation of mixed reality layouts leveraging virtual-physical semantic connections,'' in \emph{The 34th Annual ACM Symposium on User Interface Software and Technology}, 2021, pp. 282--297.

\bibitem{AgentPlacement}
Y.~Lang, W.~Liang, and L.-F. Yu, ``Virtual agent positioning driven by scene semantics in mixed reality,'' in \emph{2019 IEEE Conference on Virtual Reality and 3D User Interfaces (VR)}.\hskip 1em plus 0.5em minus 0.4em\relax IEEE, 2019, pp. 767--775.

\bibitem{transformers}
A.~Vaswani, N.~Shazeer, N.~Parmar, J.~Uszkoreit, L.~Jones, A.~N. Gomez, L.~Kaiser, and I.~Polosukhin, ``Attention is all you need,'' 2023.

\bibitem{clip}
A.~Radford, J.~W. Kim, C.~Hallacy, A.~Ramesh, G.~Goh, S.~Agarwal, G.~Sastry, A.~Askell, P.~Mishkin, J.~Clark, G.~Krueger, and I.~Sutskever, ``Learning transferable visual models from natural language supervision,'' 2021.

\bibitem{sam}
A.~Kirillov, E.~Mintun, N.~Ravi, H.~Mao, C.~Rolland, L.~Gustafson, T.~Xiao, S.~Whitehead, A.~C. Berg, W.-Y. Lo, P.~Dollár, and R.~Girshick, ``Segment anything,'' 2023.

\bibitem{cliptd}
F.~Odom, ``clip-text-decoder,'' \url{https://github.com/fkodom/clip-text-decoder}, 2022.

\bibitem{POS}
A.~Akbik, D.~Blythe, and R.~Vollgraf, ``Contextual string embeddings for sequence labeling,'' in \emph{{COLING} 2018, 27th International Conference on Computational Linguistics}, 2018, pp. 1638--1649.

\bibitem{vilt}
W.~Kim, B.~Son, and I.~Kim, ``Vilt: Vision-and-language transformer without convolution or region supervision,'' in \emph{International Conference on Machine Learning}.\hskip 1em plus 0.5em minus 0.4em\relax PMLR, 2021, pp. 5583--5594.

\bibitem{clipseg}
T.~Lüddecke and A.~S. Ecker, ``Image segmentation using text and image prompts,'' 2022.

\bibitem{gdino}
S.~Liu, Z.~Zeng, T.~Ren, F.~Li, H.~Zhang, J.~Yang, C.~Li, J.~Yang, H.~Su, J.~Zhu, and L.~Zhang, ``Grounding dino: Marrying dino with grounded pre-training for open-set object detection,'' 2023.

\bibitem{ram++}
X.~Huang, Y.-J. Huang, Y.~Zhang, W.~Tian, R.~Feng, Y.~Zhang, Y.~Xie, Y.~Li, and L.~Zhang, ``Open-set image tagging with multi-grained text supervision,'' 2023.

\bibitem{gpt4}
OpenAI, ``Gpt-4 technical report,'' 2023.

\bibitem{llava15}
H.~Liu, C.~Li, Y.~Li, and Y.~J. Lee, ``Improved baselines with visual instruction tuning,'' 2023.

\bibitem{wei2023chainofthought}
J.~Wei, X.~Wang, D.~Schuurmans, M.~Bosma, B.~Ichter, F.~Xia, E.~Chi, Q.~Le, and D.~Zhou, ``Chain-of-thought prompting elicits reasoning in large language models,'' 2023.

\bibitem{lampinen-etal-2022-language}
\BIBentryALTinterwordspacing
A.~Lampinen, I.~Dasgupta, S.~Chan, K.~Mathewson, M.~Tessler, A.~Creswell, J.~McClelland, J.~Wang, and F.~Hill, ``Can language models learn from explanations in context?'' in \emph{Findings of the Association for Computational Linguistics: EMNLP 2022}, Y.~Goldberg, Z.~Kozareva, and Y.~Zhang, Eds.\hskip 1em plus 0.5em minus 0.4em\relax Abu Dhabi, United Arab Emirates: Association for Computational Linguistics, Dec. 2022, pp. 537--563. [Online]. Available: \url{https://aclanthology.org/2022.findings-emnlp.38}
\BIBentrySTDinterwordspacing

\bibitem{openai2023gpt4}
OpenAI, ``Gpt-4 technical report,'' 2023.

\bibitem{liu2023improved}
H.~Liu, C.~Li, Y.~Li, and Y.~J. Lee, ``Improved baselines with visual instruction tuning,'' 2023.

\bibitem{gsam}
IDEA-Research, ``Grounded-segment-anything,'' \url{https://github.com/IDEA-Research/Grounded-Segment-Anything}, 2023.

\bibitem{brooks2023instructpix2pix}
T.~Brooks, A.~Holynski, and A.~A. Efros, ``Instructpix2pix: Learning to follow image editing instructions,'' 2023.

\bibitem{nyudepth}
N.~Silberman, D.~Hoiem, P.~Kohli, and R.~Fergus, ``Indoor segmentation and support inference from rgbd images.'' \emph{ECCV (5)}, vol. 7576, pp. 746--760, 2012.

\bibitem{sun3d}
J.~Xiao, A.~Owens, and A.~Torralba, ``Sun3d: A database of big spaces reconstructed using sfm and object labels,'' in \emph{Proceedings of the IEEE international conference on computer vision}, 2013, pp. 1625--1632.

\bibitem{sbert}
N.~Reimers and I.~Gurevych, ``Sentence-bert: Sentence embeddings using siamese bert-networks,'' 2019.

\bibitem{bert}
J.~Devlin, M.-W. Chang, K.~Lee, and K.~Toutanova, ``Bert: Pre-training of deep bidirectional transformers for language understanding,'' 2019.

\bibitem{arkit}
Apple, ``Arkit,'' \url{https://developer.apple.com/augmented-reality/arkit/}.

\bibitem{scenekit}
Apple., ``Scenekit,'' \url{https://developer.apple.com/documentation/scenekit}.

\bibitem{shapE}
H.~Jun and A.~Nichol, ``Shap-e: Generating conditional 3d implicit functions,'' 2023.

\end{thebibliography}

\end{document}